\pdfoutput=1

\documentclass[11pt]{article}

\usepackage[final]{coling}

\usepackage{times}
\usepackage{latexsym}

\usepackage[T1]{fontenc}

\usepackage[utf8]{inputenc}

\usepackage{microtype}

\usepackage{inconsolata}

\usepackage{graphicx}
\usepackage{booktabs}
\usepackage{multirow}
\usepackage{bbm}

\title{Large Language Models as an Indirect Reasoner: Contrapositive and Contradiction for Automated Reasoning}

\author{
 \textbf{Yanfang Zhang\textsuperscript{1}},
 \textbf{Yiliu Sun\textsuperscript{1}},
 \textbf{Yibing Zhan\textsuperscript{2}},
 \textbf{Dapeng Tao\textsuperscript{3}},
 \textbf{Dacheng Tao\textsuperscript{4}},
 \textbf{Chen Gong\textsuperscript{5}}
\\
\\
 \textsuperscript{1}Nanjing University of Science and Technology,
 \textsuperscript{2}JD Explore Academy,
 \textsuperscript{3}Yunnan University,
 \\
 \textsuperscript{4}Nanyang Technological University,
 \textsuperscript{5}Shanghai Jiao Tong University
\\
 \small{
   \textbf{Correspondence:} \href{chen.gong@sjtu.edu.cn}{chen.gong@sjtu.edu.cn}
 }
}

\begin{document}
\maketitle
\begin{abstract}
Recently, increasing attention has been focused on improving the ability of Large Language Models (LLMs) to perform complex reasoning. Advanced methods, such as Chain-of-Thought (CoT) and its variants, are found to enhance their reasoning skills by designing suitable prompts or breaking down complex problems into more manageable sub-problems. However, little concentration has been put on exploring the reasoning process, \textit{i.e.}, we discovered that most methods resort to Direct Reasoning (DR) and disregard Indirect Reasoning (IR). This can make LLMs difficult to solve IR tasks, which are often encountered in the real world. To address this issue, we propose a Direct-Indirect Reasoning (DIR) method, which considers DR and IR as multiple parallel reasoning paths that are merged to derive the final answer. We stimulate LLMs to implement IR by crafting prompt templates incorporating the principles of contrapositive and contradiction. These templates trigger LLMs to assume the negation of the conclusion as true, combine it with the premises to deduce a conclusion, and utilize the logical equivalence of the contrapositive to enhance their comprehension of the rules used in the reasoning process. Our DIR method is simple yet effective and can be straightforwardly integrated with existing variants of CoT methods. Experimental results on four datasets related to logical reasoning and mathematic proof demonstrate that our DIR method, when combined with various baseline methods, significantly outperforms all the original methods. 
\end{abstract}

\section{Introduction}
Recently, pre-trained Large Language Models (LLMs)~\cite{wang2022self,chowdhery2023palm,dubey2024llama} have shown great success in various tasks related to language comprehension~\cite{touvron2023llama,nam2024using}, content generation~\cite{agossah2023llm,liu2023use,dai2024neural}, and logical reasoning~\cite{kojima2022large,wei2022chain,dubey2024llama} due to their remarkable ability to infer from the context in zero-shot or few-shot way. To enhance the reasoning ability of LLMs, CoT~\cite{wei2022chain} encourages LLMs to explain their reasoning processes by appending some intermediate steps required to reach the answer in the prompt. Besides CoT, there are other approaches using prompts to help elicit the reasoning ability of LLMs to better solve the reasoning problems, such as Self-Consistency~\cite{wang2022self} and Least-to-Most~\cite{zhou2022least}. 

Note that most of the above mentioned methods simply perform Direct Reasoning (DR), which involves constructing logical chains from premises to the final result. However, many problems can hardly be proven or reasoned via DR. Therefore, when encountering a problem that is difficult to reach a conclusion through DR, a question arises: \textit{whether it is possible to solve the problem by performing other reasoning strategies, such as Indirect Reasoning (IR) that is logically equivalent to DR?} IR, including contrapositive and contradiction in this paper, is logically equivalent to DR, and they have been well defined in the science of logic~\cite{jourdan2016analysis}.

\begin{figure*}[t]
\begin{center}
\includegraphics[width=\textwidth]{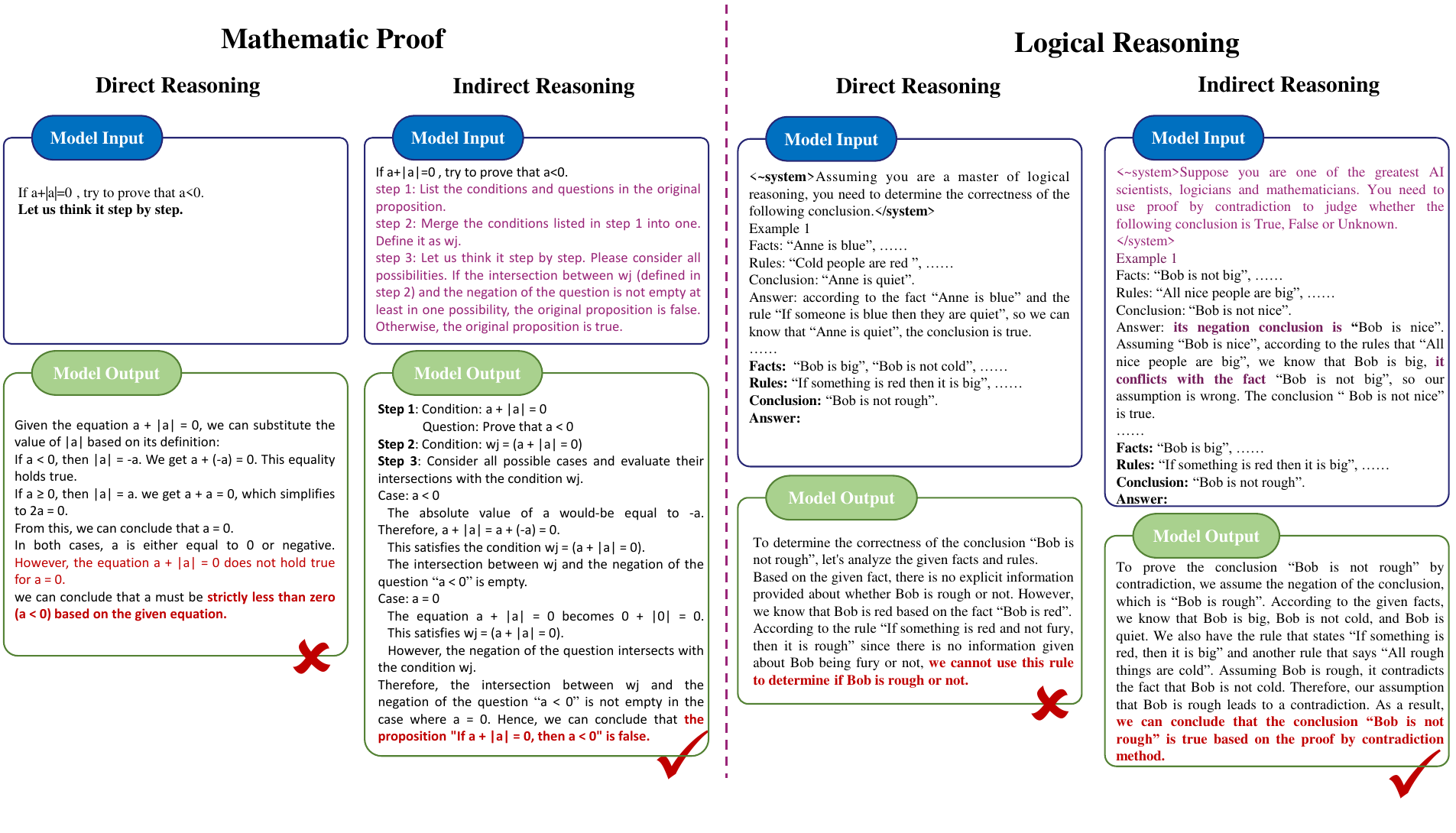}
\caption{Examples of IR using LLMs for some complex problems regarding mathematic proof and logical reasoning. Existing DR methods failed when dealing with these problems. In contrast, our method guides LLMs to use the logics of contrapositive and contradiction, resulting in accurate reasoning and successful deduction to correct answers.}
\label{fig:fig1}
\end{center}
\end{figure*}
To stimulate LLMs with IR, we devise special prompts for contradiction and contrapositive by designing instructions or examples within the intermediate reasoning process as shown in Figure~\ref{fig:fig1}. As a result, the proposed approach can induce effective IR and enhance the ability of LLMs to tackle complex reasoning tasks. Note that, IR is embarrassingly simple and general that can be directly combined with DR. Therefore, we propose Direct-Indirect Reasoning (termed ``DIR”) to further improve the reasoning ability of LLMs. Without loss of generality, we merge the results of DR and IR using the most common voting methods. Moreover, DIR can be seamlessly integrated with existing variants of CoT based on any foundation model. 

To assess the effectiveness of our DIR method, we conducted extensive experiments on two popular tasks: logical reasoning and mathematic proof, by using various LLMs as foundation models. The results indicate that our proposed method is quite effective in inspiring LLMs to achieve IR. For example, DIR has shown a noteworthy improvement of over 10.0\% in terms of the accuracy of reasoning processes of mathematic proof task. Additionally, in the logical reasoning task, it consistently outperforms various baselines and LLMs, particularly on the data that DR struggles with. The improvement is quite impressive, with a 33.4\% increase in accuracy. In particular, experimental analyses have demonstrated that the utilization of IR can aid LLMs in resolving some tasks that are arduous to accomplish through the use of DR. This enriches the reasoning paths of LLMs and improves their overall reasoning ability. Our main contributions are summarized below: 

\begin{itemize}
\item We introduce the IR strategy, including contrapositive and contradiction, into the reasoning process of LLMs. 
\item We devise a series of prompt templates that effectively stimulate LLMs to implement IR. We further introduce the DIR method, which combines DR and IR to enhance the reasoning ability of LLMs.
\item Experimental results indicate that our proposed DIR method can be effectively combined with the variants of CoT. These combined methods have shown significant performance improvement across four logical reasoning and mathematic proof benchmark datasets on three different baseline LLMs. Additionally, our method has demonstrated impressive performance in inspiring diverse reasoning chains and solving complex problems that can hardly be solved by DR.
\end{itemize}
\section{Motivation and Problem Formulation}

\begin{figure}[t]
  \centering
  \includegraphics[width=0.48\textwidth]{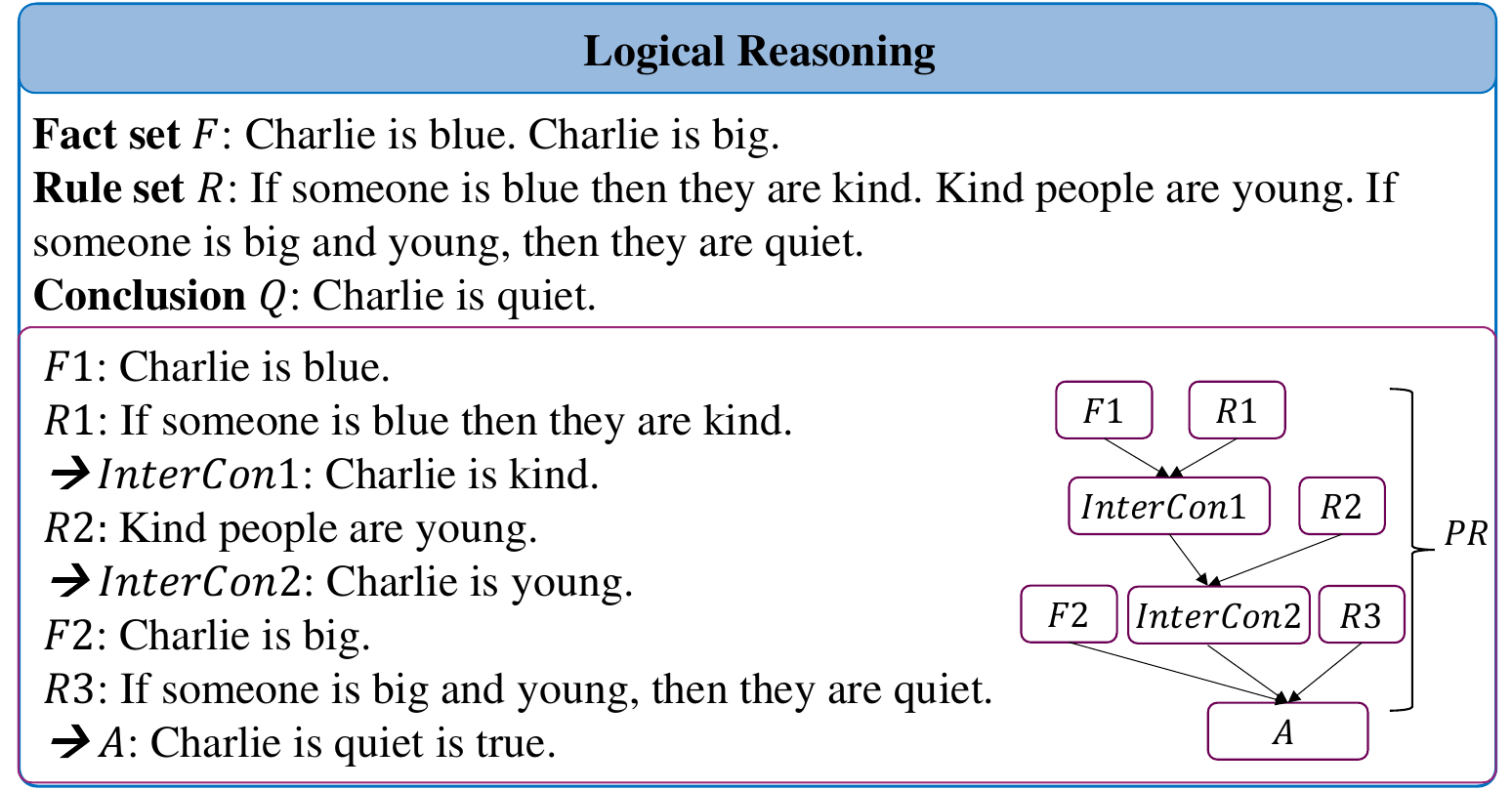}
  \caption{The illustration of some key notions in logical reasoning. The data is from ProofWriter dataset. Here $P$, $F$, $R$, $A$, $PR$, $Q$, $InterCon$ denote premise, fact, rule, answer, reasoning process, conclusion, and intermediate conclusion, respectively. The illustration of mathematic proving problem is put to appendix due to space limitation.}
  \label{fig:fig2}
\end{figure}

LLMs have shown strong ability to conduct logical reasoning in natural language. The aim of reasoning is to assess the answer $A$ to a candidate conclusion or question $Q$, and also present the reasoning process $PR$ from premises $P$ which include fact set $F$ and rule set $R$~\cite{tafjord2021proofwriter}. All the premises and conclusions are expressed in natural language. Figure~\ref{fig:fig2} shows the general illustration of logical reasoning. Mathematic proving problems are similar to logical reasoning. However, it is worth noting that it only gives fact set $F$ and question $Q$, and the rule set $R$ is usually set to prior knowledge, which means that we cannot know what rules to use in advance.

We notice that LLMs may encounter challenges with IR tasks, despite being proficient at DR, as illustrated in Figure~\ref{fig:fig1}. To further understand this phenomenon, we analyzed it by investigating whether LLMs tend to use DR in solving problems. To this end, we conducted a preliminary experiment, where 70 questions are randomly selected from each of the four datasets (\textit{i.e.}, ProofWriter, LogiQA, ProofNet, and ProofMath datasets). In these experiments, we prompt LLMs with “Let’s think step by step” and calculate the proportion of DR and IR implemented by LLMs. 

\begin{figure}[t]
  \centering
  \includegraphics[width=0.48\textwidth]{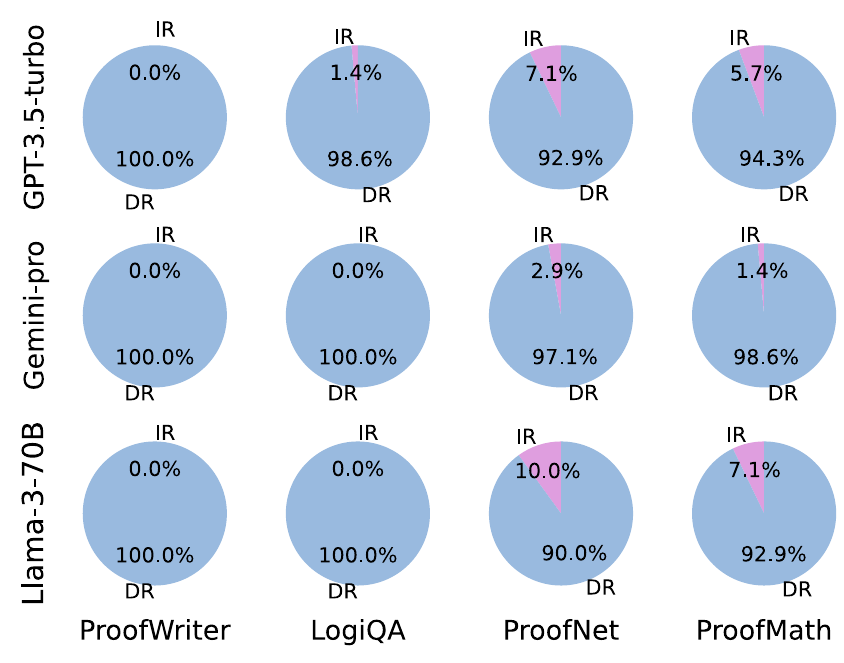}
  \caption{The proportion of IR and DR implementations deployed by LLMs on various datasets.}
  \label{fig:fig3}
\end{figure}

According to Figure~\ref{fig:fig3}, we see that LLMs rarely use IR on logical reasoning tasks, even when IR would be more appropriate. They still prefer to use DR to solve mathematic problems. Meanwhile, to our best knowledge, currently there are no relevant works explicitly performing IR. Therefore, we propose to stimulate LLMs to implement IR more effectively, which could improve their overall reasoning ability.

\section{Methodology}
\begin{figure*}[t]
\begin{center}
\includegraphics[width=\textwidth]{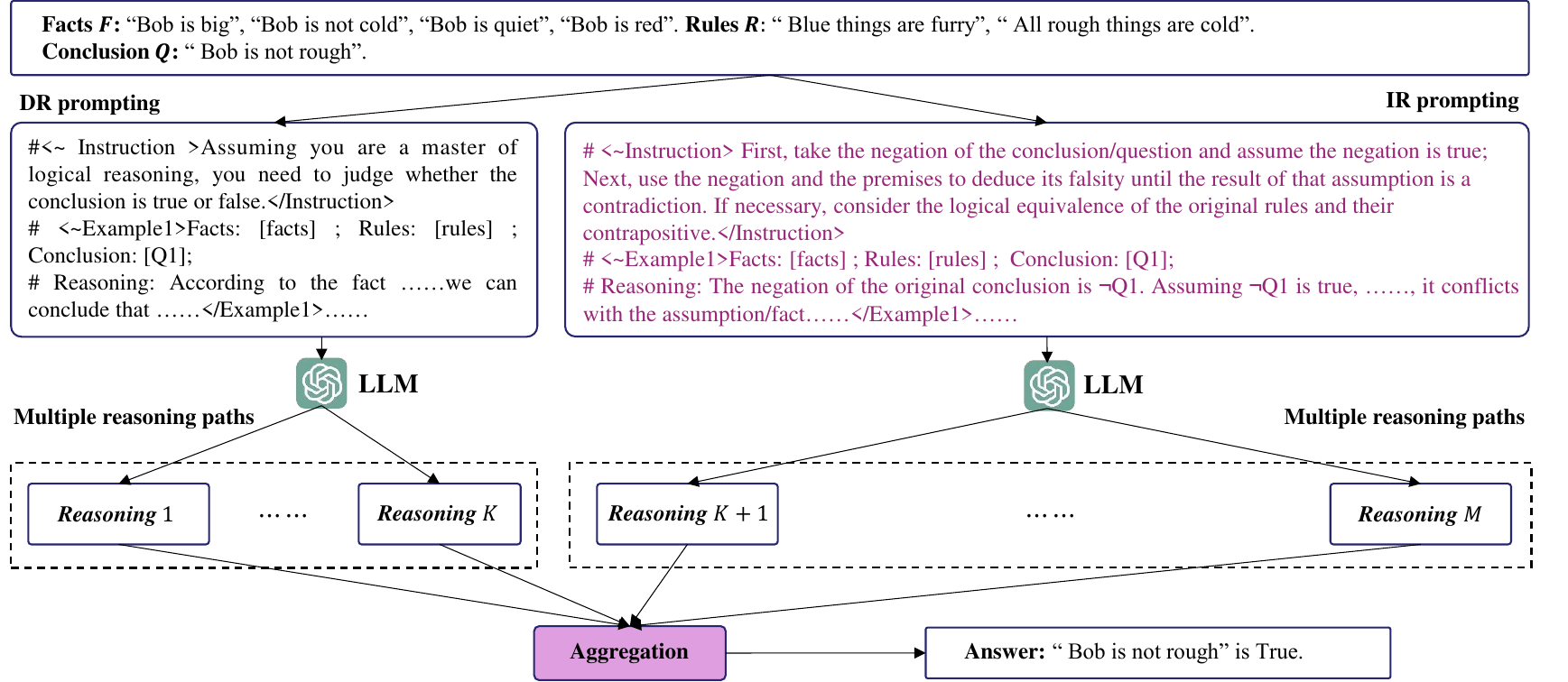}
\caption{Framework of our proposed DIR method which incorporates both DR and IR to enable multipath reasoning. Our approach involves the IR component, which stimulates LLMs through a set of crafted prompt templates that incorporate the thought of contradiction and contrapositive. Ultimately, the outcomes of both DR and IR are consolidated through majority voting.}
\label{fig:fig5}
\end{center}
\end{figure*}

In this section, we present a comprehensive overview of our DIR approach as shown in Figure~\ref{fig:fig5}. Our method begins with an introduction to the principles of IR, which include contradiction and contrapositive (Section~\ref{sec:sec41}). Then we outline a method for guiding LLMs in the application of IR by devising prompt templates that implement the reasoning process of contradiction and contrapositive (Section~\ref{sec:sec42}). Lastly, we provide a detailed description of the combination method for DR and IR in Section~\ref{sec:sec43}.

\subsection{Contrapositive and Contradiction} \label{sec:sec41}
In mathematics and some practical applications, there are circumstances where direct proof may not be feasible or effective. In such cases, the methods of indirect proof are often used to verify a statement. There are two popular methods for indirect proof, which are: contrapositive method and contradiction method. Next, we will explain these two methods in detail.

\textbf{Contrapositive.} It is based on the fact that an implication is equivalent to its contrapositive, namely:
\begin{equation}
    p\rightarrow q\Leftrightarrow \neg p\vee q,
\end{equation}
\begin{equation}
    \neg q\rightarrow \neg p\Leftrightarrow \neg (\neg q)\vee \neg p\Leftrightarrow q\vee \neg p.
\end{equation}
According to the commutative law, one can have:
\begin{equation}
    p\rightarrow q\Leftrightarrow \neg q\rightarrow \neg p.
\end{equation}
Therefore, when we get a fact ``If $p$, then $q$”, we can also know that if $\neg q$ then $\neg p$.

\textbf{Contradiction.} The world-renowned mathematician G. H. Hardy called proof-by-contradiction ``one of a mathematician's finest weapons”. Actually, this method has been widely used in mathematics, logic, and philosophy to establish the validity of various statements and arguments. Proof-by-contradiction involves the original statement and its negation. These two statements are opposites to each other, meaning that if the original statement is true, the negation of the original statement is false; and if the original statement is false, the negation of the original statement is true. Therefore, we consider a reasoning equivalence as:
\begin{equation}
    \neg (p\rightarrow q)\Leftrightarrow \neg [(\neg p)\vee q]\Leftrightarrow p\wedge (\neg q).
\end{equation}

\subsection{Indirect Reasoning}\label{sec:sec42}
In the principles of IR described in the previous section, we inspire LLMs to implement IR by designing appropriate prompt templates as shown in Figure~\ref{fig:fig5}. To implement contradiction, the entire reasoning process is involved. First, we take the negation of the conclusion and assume it to be true. Subsequently, we deduce the negation along with the premises until a conflict arises. Finally, we conclude that the negation of the conclusion is false, and therefore, the original conclusion must be true. In addition, as depicted in Figure~\ref{fig:fig2}, certain rules are employed during the reasoning process. Based on the principle of contrapositive discussed earlier, we can deduce that the contrapositive of these rules and their original rules are logically equivalent. The contrapositive can assist LLMs in enhancing their comprehension of rules and their ability to apply them efficiently. For instance, when presented with the fact ``Bob does not drive to work” and the rule ``If the weather is fine, Bob drives to work”, humans can apply the equivalence of contrapositive to deduce that the rule is equivalent to ``If Bob does not drive to work, the weather is not fine”. This allows them to conclude that ``The weather is not fine” based on the rule. In the following, we present relevant instructions and examples with IR processes to achieve contradiction and contrapositive.

\textbf{Zero-shot Scenario.} We implement a contradiction by following instructions: ``First, take the negation of the conclusion/question and assume the negation is true; Next, use the negation and the premises to deduce its falsity until the result of that assumption is a contradiction”. Also for contrapositive, LLMs are prompted using ``If necessary, consider the logical equivalence of the original rules and their contrapositive”.

\textbf{Few-shot Scenario.} In addition to the above instructions, the examples with intermediate reasoning steps incorporating contradiction and contrapositive also facilitate LLMs to implement IR. To facilitate the effective implementation of IR for LLMs, we craft a set of prompt templates that incorporate the concepts of contradiction and contrapositive into the reasoning process (see Appendix~\ref{apx:C}). 

\subsection{Direct-Indirect Reasoning}\label{sec:sec43}
From the above description, it can be inferred that the proposed IR method can be directly combined with DR in existing methods, such as SC~\cite{wang2022self},  CR~\cite{Cumulative2023} and MulAD~\cite{duimproving}. Therefore, we propose a DIR method by combining IR and DR. This will enrich the reasoning paths to solve complex problems. It can be seen that IR is a straightforward approach that involves negating the conclusion and treating the negation as a premise. That is to say, IR does not impose any additional constraints on the reasoning process. Therefore, the proposed DIR method can be easily incorporated into existing CoT variants to improve reasoning proficiency. 

Various techniques exist for aggregating the results of multipath reasoning. One straightforward way is to select the most commonly occurring results, while another involves utilizing the log probability of the output of LLMs. In this paper, we utilize voting to select the most frequently occurring results. We sample $M$ candidate reasoning chains from LLMs and $\left \{  A_{i}\right \} _{i=1}^{M}$ is the set of answers generated from these chains. Let $\mathcal{A}=\left \{ \widehat{A}_{s}\right\} _{s=1}^{\left | \mathcal{A}  \right | }$ be the set of all possible answers for the question. We then select the answer from $\mathcal{A}$ with the highest probability $P(\widehat{A}_{s})$ and $P(\widehat{A}_{s})$  can be formulated as below:
\begin{equation}
    P(\widehat{A}_{s})=\frac{1}{M} {\textstyle \sum_{i=1}^{M}\mathbbm{I}(A_{i}=\widehat{A} _{s})},
\end{equation}
where $\mathbbm{I}(\cdot)$ is the indicator function. 
\section{Empirical Study}
\subsection{Setup}
To evaluate the effectiveness of our proposed DIR method, we apply our method to four typical reasoning datasets, namely ProofWriter, LogiQA, ProofNet, and ProofMath. Here we choose the popular CoT-based prompt methods to implement both DR and IR, which are: 
\begin{itemize}
\item CoT~\cite{wei2022chain}. It guides LLMs to reason by utilizing a few examples containing reasoning process.
\item SC~\cite{wang2022self}. It samples multiple reasoning chains and selects the final result by voting.
\item CR~\cite{Cumulative2023}. It is similar to ToT~\cite{yao2024tree} and solves problems using a thought search tree. However, CR stores all the historically correct intermediate thoughts. 
\item MulAD~\cite{duimproving}. It employs multiple agents, each powered by an LLM, to propose and debate their individual reasoning processes over multiple rounds to arrive at a common final answer.
\end{itemize}
In subsequent experiments, the proposed DIR algorithms are configured to have the same number of reasoning candidates sampled from LLMs as their corresponding baseline algorithms. This configuration can make the computational complexity to be consistent among DIR and baseline methods in the experiment. Additionally, the number of reasoning candidates sampled from LLMs for DR and IR in DIR is set to be equal. Specifically, for LogiQA and ProofWriter datasets, we set the number of reasoning candidates to 16, and for ProofNet and ProofMath datasets, it is set to 4. For more detailed parameter settings in the implementation please refer to Appendix~\ref{apx:A} and the designed prompt templates for the experiment are available in Appendix~\ref{apx:C}.
\subsection{Evaluation Metrics}
The evaluation of reasoning performance for a method includes the correctness investigation on the answer and the reasoning process. Therefore, here we use three metrics, namely accuracy of answer (AA), accuracy of reasoning processes (AP), and diversity of reasoning processes (DP). We use these three indicators to comprehensively evaluate the quality of reasoning or proof. The definitions of AA, AP, and DP are:
\begin{equation}
    AA=\frac{AN}{N},AP=\frac{PN}{N},DP=\frac{1}{N}\sum d_{i},
\end{equation}
where $N$ is the number of examples in the test set; $AN$ and $PN$ are the numbers of examples with correct answer prediction and correct reasoning process prediction, respectively; and $d_{i}$ is the number of correct reasoning methods for the $i$-th example.
\subsection{Main Results}
\setlength{\tabcolsep}{3pt}
\begin{table}[t]

    \centering
    \footnotesize
    \begin{tabular}{lccc}
         \toprule
            & GPT-3.5-turbo & Gemini-pro & Llama-3-70B\\ \midrule
        CoT     &  28.5\% &  26.3\% &26.8\%\\ \midrule
        SC     & 29.1\%  &  27.4\%  &29.6\%\\
        SC-DIR     & \textbf{36.3\%}  &  \textbf{31.3\%}&\textbf{34.6\%} \\ \midrule
        CR     & 25.1\%  &  22.3\% &22.9\%\\
    CR-DIR    & \textbf{29.1\%} & \textbf{25.7\%} &\textbf{26.8\%}\\ \midrule
    MulAD     & 31.3\%  &  26.8\%&30.2\% \\
    MulAD-DIR    & \textbf{38.0\%} & \textbf{31.8\%}&\textbf{35.8\%}\\ 
    \bottomrule
    \end{tabular}
    \caption{Reasoning accuracy of various methods on LogiQA dataset.}
    \label{tab:tab1}
\end{table}

\textbf{LogiQA.} The LogiQA~\cite{liu2021logiqa} dataset comprises 8,678 paragraph-question pairs, and each of them is accompanied by four answer choices. To assess the reasoning ability of LLMs, we conduct a thorough examination of 179 of these questions with minimal dependence on external sources. This allows us to more rigorously evaluate the logical reasoning capabilities of these models~\cite{sun2023indeterminacy}. AA is used to evaluate the accuracy of reasoning in this task.

The results of reasoning on LogiQA are presented in Table~\ref{tab:tab1}. The results indicate that integrating SC, CR and MulAD with DIR leads to a consistent improvement in accuracy. In the GPT-3.5-turbo scenario, DIR outperforms SC by 7.2\%. This improvement is mainly due to the fact that IR can offer more diverse reasoning paths. Detailed case descriptions can be found in Appendix~\ref{apx:B}.

\begin{table}[t]
    \centering
    \footnotesize
    \begin{tabular}{lccc}
         \toprule
            & GPT-3.5-turbo & Gemini-pro & Llama-3-70B\\ \midrule
        CoT     &  48.5\% &  42.0\%  & 71.0\%\\ \midrule
        SC     & 53.0\%  &  46.0\% & 74.5\% \\
        SC-DIR     & \textbf{55.5\%}  &  \textbf{57.0\%} & \textbf{82.5\%}\\ \midrule
        CR     & 42.5\%  &  36.5\% & 61.5\%\\
    CR-DIR    & \textbf{45.5\%}  & \textbf{49.5\%} & \textbf{73.0\%}\\ \midrule
    MulAD    & 54.0\%  &  48.5\% & 78.0\%\\
    MulAD-DIR    & \textbf{58.5\%}  & \textbf{59.5\%} & \textbf{87.5\%} \\ 
    \bottomrule
    \end{tabular}
    \caption{Reasoning accuracy of various methods on ProofWriter dataset.}
    \label{tab:tab2}

\end{table}

\textbf{ProofWriter.} ProofWriter~\cite{tafjord2021proofwriter} dataset is a widely used benchmark dataset regarding logical reasoning. We utilize the OWA subset of ProofWriter, which is categorized into five subsets based on the number of hops required in the reasoning. For our purposes, we choose the 5-hop subset, which consists of questions requiring 0, 1, 2, 3, and 5 hops. Following the guidelines outlined in~\cite{kazemi2023lambada}, we randomly select 200 questions from this subset for testing.

The results in Table~\ref{tab:tab2} suggest that the implementation of DIR can significantly enhance the reasoning performance of all baseline methods, with a performance improvement more than 10.0\% on Gemini-pro. Through an analysis of the reasoning process, it is discovered that IR enhances the reasoning ability of LLMs by prompting them to explore more reasoning chains. This is shown in detail in the Case Study in Section~\ref{sec:sec44}.

\begin{table}[t]
    \centering
    \footnotesize
    \begin{tabular}{lcccccc}
         \toprule
           &  \multicolumn{2}{c}{GPT-3.5-turbo} & \multicolumn{2}{c}{Gemini-pro} & \multicolumn{2}{c}{Llama-3-70B} \\ \cmidrule{2-7}
        & AP & DP & AP & DP & AP & DP \\ \midrule
        CoT     &  46.0\% &  0.46 &  44.0\%& 0.44 &  40.0\%& 0.40 \\ \midrule
        SC     & 72.0\% & 1.28 &  60.0\% & 1.00 &  66.0\% & 1.02\\
        SC-DIR     & \textbf{82.0\% } & \textbf{1.88} &  \textbf{72.0\% } & \textbf{1.52}&  \textbf{76.0\% } & \textbf{1.48} \\ \midrule
        CR     & 64.0\% & 1.06 &  50.0\% & 0.82 &  58.0\% & 0.84 \\
    CR-DIR    & \textbf{76.0\% } & \textbf{1.60} & \textbf{62.0\% } & \textbf{1.24} & \textbf{66.0\% } & \textbf{1.04}\\ \midrule
    MulAD     & 74.0\% & 1.26 &  62.0\% & 1.04 &  64.0\% & 1.14 \\
    MulAD-DIR    & \textbf{84.0\% } & \textbf{1.64} & \textbf{72.0\% } & \textbf{1.38} & \textbf{78.0\% } & \textbf{1.44}\\ 
    \bottomrule
    \end{tabular}
    \caption{Reasoning accuracy of various methods on ProofNet dataset.}
    \label{tab:tab3}
\end{table}

\textbf{ProofNet.} The ProofNet dataset~\cite{azerbayev2023proofnet} is a collection of problems used for assessing the ability of automated systems to formalize and verify mathematic proofs at an undergraduate level. To evaluate the accuracy of LLMs in mathematic proof task, we use two metrics, namely AP and DP. The evaluation of the process is entrusted to undergraduate math majors. As a cost-efficient approach, we opt to randomly select 50 questions from ProofNet for testing purposes.

The findings shown in Table~\ref{tab:tab3} prove that the DIR, when used in conjunction with SC, CR and MulAD, is better than the original methods, with a maximum improvement of 14.0\%. At the same time, the findings disclosed by DP indicate that DIR successfully motivates LLMs to produce various reasoning chains, indicating its effectiveness. 

\begin{table}[t]
    \centering
    \footnotesize
    \begin{tabular}{lcccccc}
         \toprule
           &  \multicolumn{2}{c}{GPT-3.5-turbo} & \multicolumn{2}{c}{Gemini-pro} & \multicolumn{2}{c}{Llama-3-70B}\\ \cmidrule{2-7}
        & AP & DP & AP & DP & AP & DP\\ \midrule
        CoT     &  56.0\% &  0.56 &  47.0\%& 0.47&  51.0\%& 0.51\\ \midrule
        SC     & 63.0\% & 0.65 &  59.0\% & 0.62&  62.0\% & 0.64 \\
        SC-DIR     & \textbf{77.0\%} & \textbf{0.88} &  \textbf{71.0\% } & \textbf{0.76}&  \textbf{73.0\% } & \textbf{0.75} \\ \midrule
        CR     & 57.0\% & 0.59 &  54.0\% & 0.56&  58.0\% & 0.60 \\
    CR-DIR    & \textbf{72.0\% } & \textbf{0.77} & \textbf{68.0\% } & \textbf{0.71}& \textbf{67.0\% } & \textbf{0.71}\\ \midrule
    MulAD     & 63.0\% & 0.64 &  58.0\% & 0.59&  63.0\% & 0.65 \\
    MulAD-DIR    & \textbf{78.0\% } & \textbf{0.84} & \textbf{71.0\% } & \textbf{0.73}& \textbf{75.0\% } & \textbf{0.77}\\
    \bottomrule
    \end{tabular}
    \caption{Reasoning accuracy of various methods on ProofMath dataset.}
    \label{tab:tab4}
\end{table}

\textbf{ProofMath.} As revealed by~\cite{yang2024leandojo}, the above ProofNet is publicly available on GitHub before the data of LLMs used in the experiment cutoff date. Therefore, there is a potential risk that LLMs are pre-trained with their standard proof. To obtain a more comprehensive and accurate evaluation, we create a new dataset called ``ProofMath”. This dataset contains 100 mathematic proof problems from junior and senior high schools. We make the dataset diverse in terms of problem difficulty (see Appendix~\ref{apx:A}) so that we can comprehensively assess the reasoning ability of the DIR techniques. Similar to ProofNet, we employ both AP and DP metrics for evaluation purposes. 

The results displayed in Table~\ref{tab:tab4} indicate that employing DIR instead of DR results in a 10.0\% enhancement in terms of AP. It is worth noting that this enhancement rises to 15.0\% in the presence of GPT-3.5-turbo. Furthermore, DP illustrates the positive influence of DIR in encouraging LLMs to explore more reasoning paths.

\subsection{Discussion}\label{sec:sec44}
\begin{table}[t]
    \centering
    \footnotesize
    \begin{tabular}{lcccc}
         \toprule
          &  & \fontsize{8pt}{8pt}\selectfont{GPT-3.5-turbo} & \fontsize{8pt}{8pt}\selectfont{Gemini-pro} & \fontsize{8pt}{8pt}\selectfont{Llama-3-70B}\\ \midrule
        \multirow{2}{*}{\fontsize{8pt}{8pt}\selectfont{ProofWriter\_S}}  &  DR   &  17.3\% &  21.3\% & 29.3\%\\
           &  IR & \textbf{50.7\%}  &  \textbf{47.3\%} &  \textbf{66.7\%} \\ \midrule
        \multirow{2}{*}{\fontsize{8pt}{8pt}\selectfont{ProofMath\_S}}  & DR   & 57.1\%  &  42.9\% & 48.6\%\\ 
          & IR   & \textbf{82.6\%}   &  \textbf{62.9\%} &  \textbf{74.3\%}\\ \bottomrule
    
    \end{tabular}
    \caption{Reasoning accuracy comparison of DR and IR on ProofWriter\_S and ProofMath\_S datasets.}
    \label{tab:tab5}
\end{table}
\textbf{IR can prompt LLMs to implement effective indirect reasonings.} In the following experiments, we thoroughly evaluate whether our proposed IR method can prompt LLMs to perform effective IR in solving IR tasks. With this in mind, we carefully select 150 data items from the ProofWriter dataset termed ``ProofWriter\_S” and 35 data items from the ProofMath dataset termed ``ProofMath\_S” to showcase the advantages of IR. We use SC as a baseline method to perform DR and IR, with four reasoning candidates sampled from LLMs. To evaluate the ability of LLMs to implement effective IR, we select AP as the evaluation metric. Table~\ref{tab:tab5} shows the performance comparison between DR and IR on these subsets. 

Based on the results, it is evident that IR significantly outperforms DR counterparts across multiple LLMs. Specifically, IR showcases enhancements of 33.4\% for ProofWriter and 25.5\% for ProofMath when using GPT-3.5-turbo. Our analysis indicates that while DR can address certain 0-hop issues within ProofWriter\_S, it fails to provide accurate reasoning for more difficult questions. In contrast, IR can solve problems of various levels of complexity by using contradiction and contrapositive techniques.

\begin{table}[t]
    \centering
    \footnotesize
    \begin{tabular}{lcccc}
         \toprule
          &  & \fontsize{8pt}{8pt}\selectfont{GPT-3.5-turbo} & \fontsize{8pt}{8pt}\selectfont{Gemini-pro} & \fontsize{8pt}{8pt}\selectfont{Llama-3-70B}\\ \midrule
        \multirow{2}{*}{\fontsize{8pt}{8pt}\selectfont{ProofWriter\_S}}  &  DR   &  15.3\% &  20.0\%  &28.7\%\\
           &  IR & \textbf{49.3\%}  &  \textbf{44.7\%}  &  \textbf{62.7\%}\\ \midrule
        \multirow{2}{*}{\fontsize{8pt}{8pt}\selectfont{ProofMath\_S}}  & DR   & 54.3\%  &  45.7\% & 48.6\%\\ 
          & IR   & \textbf{80.0\%}   &  \textbf{60.0\%} &  \textbf{68.6\%}\\ \bottomrule
    \end{tabular}
    \caption{Reasoning accuracy comparison of DR and IR by zero-shot prompts.}
    \label{tab:tab6}
\end{table}

\textbf{IR works for zero-shot prompts.} We conduct a study to determine if IR can prompt LLMs to implement IR through zero-shot prompts. To achieve this, the examples from the prompt templates are removed and only the relevant instructions are utilized, as illustrated in Table~\ref{tab:tab6}. The results reveal that LLMs can be effectively stimulated to implement IR via zero-shot prompts, resulting in significant improvements compared with DR.

\begin{figure}[t]
  \centering
  \includegraphics[width=\linewidth]{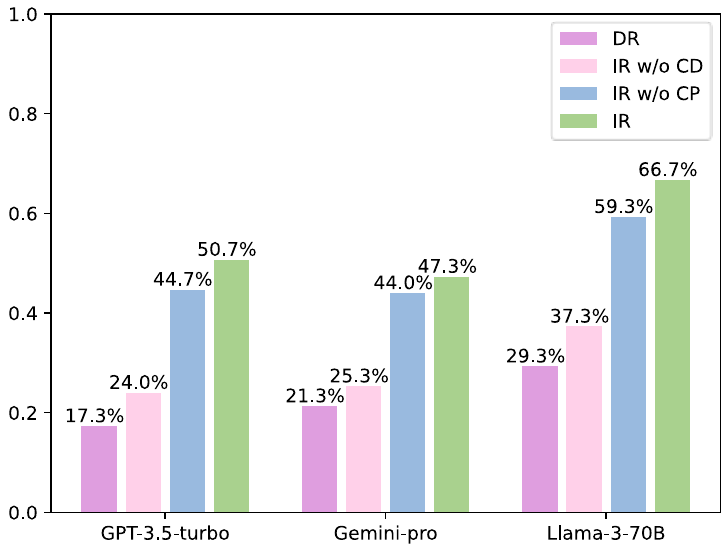}
  \caption{The impact of contrapositive and contradiction on IR.}
  \label{fig:fig7}
\end{figure}

\textbf{The impact of contrapositive and contradiction.} We conduct ablative experiments on ProofWriter\_S to assess the impact of contrapositive and contradiction on IR. In the experiment, the instructions for contrapositive and contradiction of the template are respectively removed. We utilize SC as a baseline method to perform DR and IR, employing four reasoning candidates. The results, quantified by the AP metric, are presented in Figure~\ref{fig:fig7}. The results show that removing either contrapositive (termed ``IR w/o CP”) or contradiction (termed ``IR w/o CD”) disrupts the performance of IR.

\begin{figure*}[t]
\begin{center}
\includegraphics[width=0.95\textwidth]{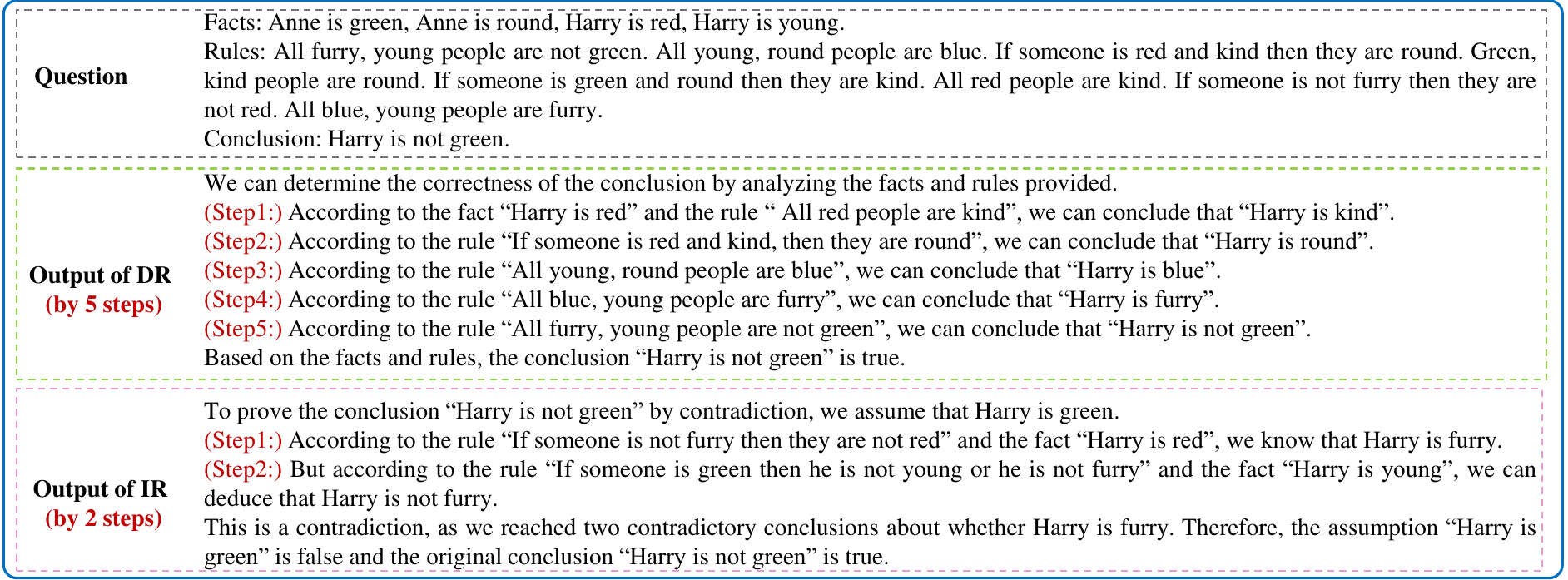}
\caption{IR uses fewer reasoning steps to reach a conclusion in logical reasoning.}
\label{fig:fig6}
\end{center}
\end{figure*}

\textbf{Case Study.}\label{sec:ct} We analyze several cases to gain a better understanding of the reasoning process of IR. We find that using IR leads to more diverse reasoning paths as shown in Figure~\ref{fig:fig6}. In some multi-step reasoning scenarios, IR helps LLMs to reach the answer with fewer steps than DR in many circumstances, thereby decreasing the likelihood of making errors in the reasoning processes of LLMs as described in~\cite{kazemi2023lambada}. More case studies are available in Appendix~\ref{apx:B}.

\section{Related Work}
Reasoning ability, as a basic ability of LLMs, has received great attention recently due to its great importance. Despite the notable improvements made by CoT~\cite{wei2022chain}, LLMs are still struggling with the tasks that require complex or high-order multi-step reasoning, such as logical reasoning and mathematic proof. Therefore, intensive research efforts have been dedicated to addressing the aforementioned issues. Generally, they can be categorized as follows.

\textbf{Fine-tuning-based methods.} These methods aim to improve the reasoning ability of LLMs through supervised fine-tuning. Usually, LLMs are fine-tuned by the samples which require manual labeling of reasoning processes, such as~\cite{ouyang2022training,wang2022super}. However, it can be labor-intensive due to the costly labeling of complex reasoning processes. The works of~\cite{shridhar2022distilling,zelikman2022star} first used LLMs to generate reasoning processes, but only the samples with correct results are selected for fine-tuning LLMs to reduce the labeling cost. Additionally, fine-tuned LLMs on specific tasks can suffer from the problem of ``catastrophic forgetting”, which means that the original knowledge inherited by the pre-trained LLMs will be lost and thus the ability to generalize to downstream tasks will be weakened. To this end, ~\newcite{cheng2023uprise} trained a prompt retriever using the output scores of LLMs. When fine-tuning, LLMs are frozen just as a data labeler which effectively reduces the impact on LLMs.

\textbf{Tool-based methods.} Tool-based methods propose to utilize external tools to augment the capabilities of LLMs in accomplishing complex tasks~\cite{qin2023toolllm, schick2024toolformer}. Moreover, ~\newcite{jin2024genegpt,yang2023chatgpt} augment LLMs with external real-time knowledge or domain-specific information through specific tools. Additionally, Retrieval-Augmented Generation (RAG) related methods~\cite{gao2023retrieval, ma2023query,peng2024graph} have received a lot of attention recently, and these methods improve the reasoning ability of LLMs by incorporating external knowledge. 

\textbf{CoT-based methods.} CoT-based methods use prompts to help elicit the reasoning ability of LLMs to better solve the reasoning problems~\cite{kojima2022large, wei2022chain, zhang2022automatic}, which is also closely related to our paper. The common CoT methods contain zero-shot CoT~\cite{kojima2022large} and few-shot CoT~\cite{wei2022chain}. Meanwhile, recent researches show that different variants of CoT can improve the reasoning ability of LLMs. For instance, the method in~\cite{zhang2022automatic} enhances the performance by optimally selecting examples in the prompt. Additionally, external information can be introduced to increase the credibility of results, as proposed in~\cite{he2022rethinking}. Some different approaches are proposed in~\cite{besta2024graph,drozdov2022compositional, yao2024tree} to decompose complex problems into smaller subproblems to enhance the reasoning ability of LLMs. Furthermore, recent developments indicate that multi-agent debates~\cite{wang2024rethinking, duimproving} can improve reasoning skills in LLMs.

However, as mentioned in the introduction, the previous researches mainly focus on DR, which will meet difficulties in some complex reasoning procedures. Therefore, our work aims to explore IR combined with DR methods to further improve the reasoning ability of LLMs.

\section{Conclusion}
In this paper, we propose a DIR method to enhance the reasoning power of LLMs by tailored prompts. IR can well compensate for problems which are not directly derivable from known facts and rules. We validate the effectiveness of the DIR method in logical reasoning and mathematic proof tasks, and the results well confirm the usefulness of the proposed IR strategy. Considering that the IR in this paper only involves the simple thoughts of contrapositive and contradiction, in the future, we can explore the possibility of integrating other more complex logical laws to make LLMs further improve their reasoning skills.

\section*{Limitations} 
Our approach has yielded consistent performance improvements across various LLMs. However, the extent of these improvements varies depending on the specific LLM. Upon analyzing the experimental outcomes, we have observed that GPT-3.5-turbo performs IR more effectively and with greater stability than Gemini-pro in most cases. These findings suggest that the foundational model has an impact on the effectiveness of IR in LLMs.
\section*{Acknowledgments}
This work was supported by the Major Science and Technology Innovation 2030 ``New Generation Artificial Intelligence” key project (No. 2021ZD0111700), NSF of China (Nos: 62336003, 12371510), and NTU RSR and Start Up Grants. 
\bibliography{coling_latex}
\newpage
\appendix
\section{Illustration of Mathematic Proof}
\begin{figure}[t]
  \centering
  \includegraphics[width=\linewidth]{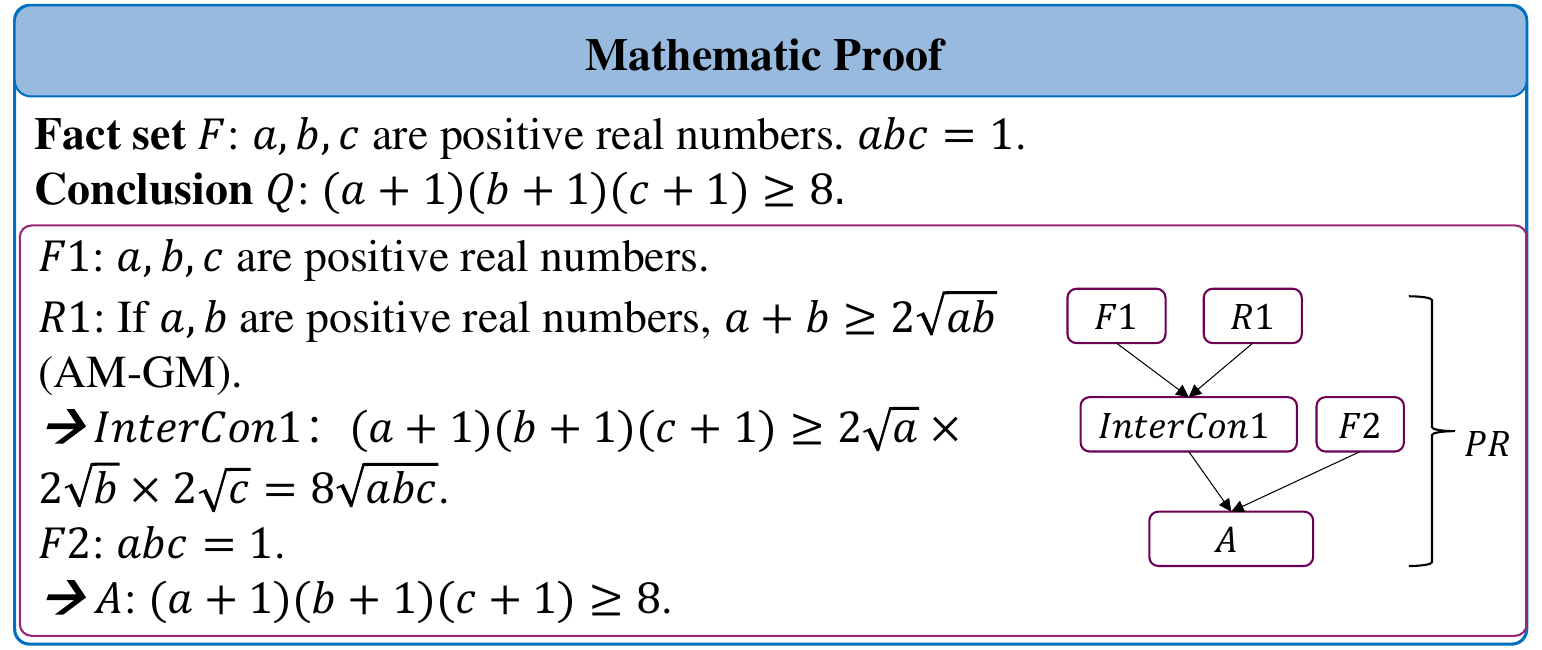}
  \caption{The illustration of some key notions in mathematic proof.}
  \label{fig:figA16}
\end{figure}

Figure~\ref{fig:figA16} shows the general illustration of mathematic proof. Similar to the logical reasoning task mentioned earlier, the goal of a mathematic proof is to prove a conclusion based on given facts and rules. However, in mathematic proof task, the rules are often not explicitly provided but are instead treated as implicit knowledge generally embedded in LLMs.

\section{Implementation Details}\label{apx:A}
Below are some detailed supplementary introductions that outline the relevant implementation details discussed in the paper.

\subsection{Parameter Settings}
In the experiment, we choose GPT-3.5-turbo, Gemini-pro and Llama-3-70B as the basic LLMs. Among them, the temperature for sampling the output of GPT-3.5-turbo is set to 0.7 as~\cite{wang2022self}, the temperature for sampling the output of Gemini-pro is set to 0.9 in Google AI Studio\footnote{\thanks{https://cloud.google.com/vertex-ai/docs/generative-ai/model-reference/gemini}} and it is set to 0.6 for Llama-3-70B in the official Llama repo\footnote{\thanks{https://llama.meta.com/docs/llama-everywhere/running-meta-llama-on-linux/}}. It is pertinent to note that the CR generates intermediate thoughts at a lower temperature when compared with the sampling temperature. To this end, the temperature is set to 0.1, 0.3 and 0.3 for GPT-3.5-turbo, Gemini-pro and Llama-3-70B, respectively. For MulAD, the number of agents is set to 3 and the number of rounds is set to 2. The prompting templates for multiple rounds of debate align with those detailed in~\cite{duimproving}.

\subsection{Details of ProofMath}
We develop ProofMath, a collection of 100 mathematic proof questions tailored for junior and senior high school students. These questions and proofs are presented in natural language. The questions have been drawn from exercises in junior and senior high school mathematics textbooks as well as on a specific website\footnote{http://1v1.zuoyebang.com/}. The selection of questions aims to encompass a wide array of topics, spanning varying degrees of complexity as shown in Table~\ref{tab:tabA2}. The questions are carefully selected by three graduate students with robust educational backgrounds in science and engineering, each possessing considerable mathematic expertise. The proofs for these questions are adapted from the textbooks or the website and subsequently verified by three experts.

\begin{table*}[t]
    \centering
    \footnotesize
    \begin{tabular}{lllc}
        \toprule
        \multicolumn{3}{c}{Scope of Mathematic Knowledge} & Quantity\\ \midrule
        \multirow{5}{*}{Junior School}     &  \multirow{2}{*}{Algebra} & Integral and Fractional Formulas  & 6\\ 
        & &  Function & 11 \\ \cmidrule{2-4}
        &  \multirow{3}{*}{Geometry} & Lines & 7\\ 
        & & Triangle & 8\\ 
        & & Polygon & 8\\ \midrule
        \multirow{5}{*}{High School}     &  \multirow{3}{*}{Algebra} & Set  & 13\\
       & & Function & 19\\ 
       & & Series & 12\\ \cmidrule{2-4}
       &  \multirow{2}{*}{Geometry} & Solid Geometry & 7\\ 
       & & Analytic Geometry & 9\\
    \bottomrule
    \end{tabular}
    \caption{The distribution of questions in ProofMath.}
    \label{tab:tabA2}

\end{table*}

\begin{figure}[t]
  \centering
  \includegraphics[width=\linewidth]{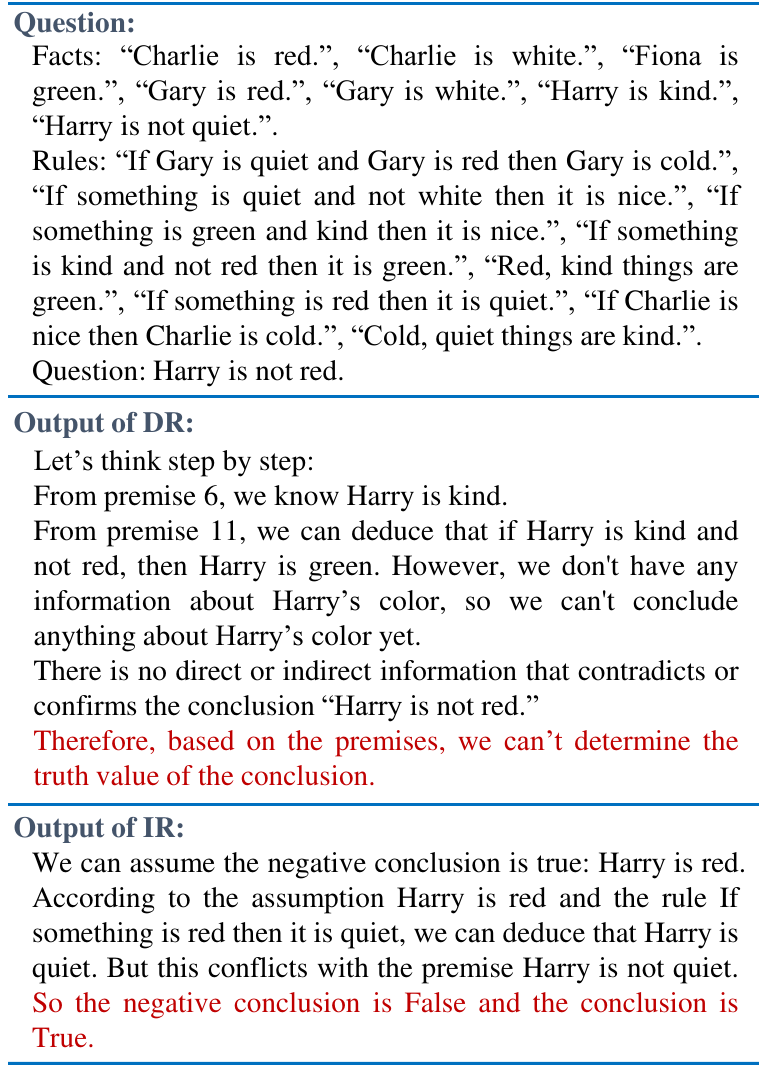}
  \caption{An example in ProofWriter demonstrates that using IR leads to accurate reasoning while using DR leads to failure.}
  \label{fig:figA11}
\end{figure}

\section{Additional Experimental Results}\label{apx:B}
\begin{figure}[t]
  \centering
  \includegraphics[width=\linewidth]{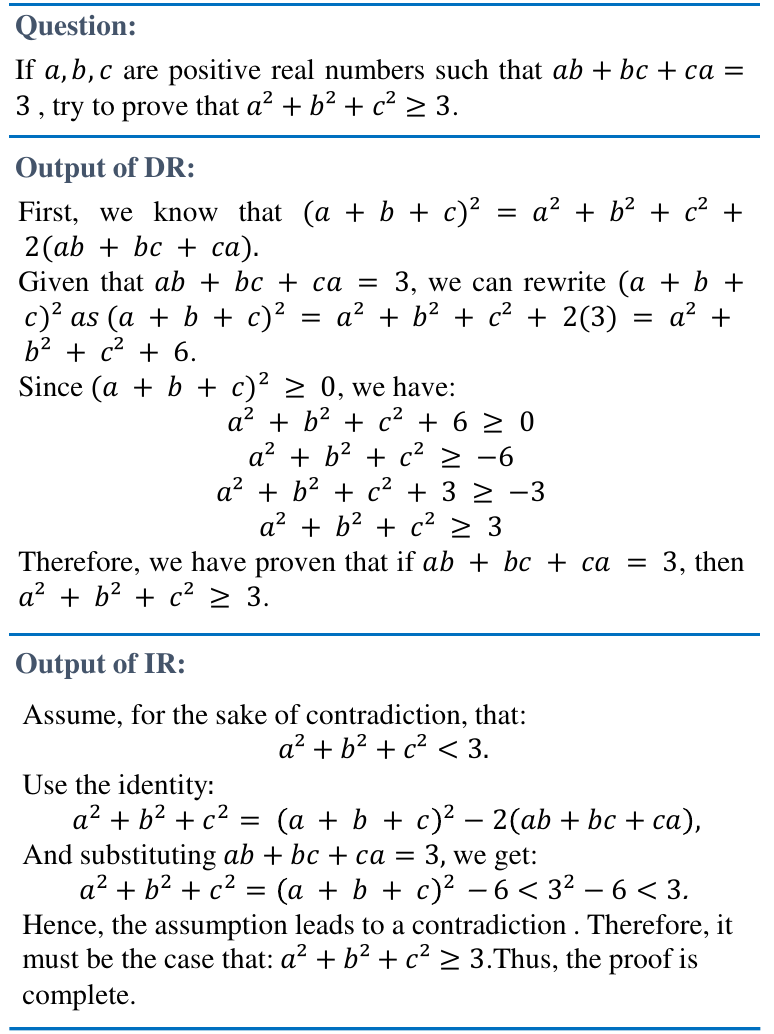}
  \caption{An example in ProofMath demonstrates that using IR leads to accurate reasoning while using DR leads to failure.}
  \label{fig:figA12}
\end{figure}

\textbf{Further analysis of the reasoning process of IR and DR.} To better understand the impact of IR on the reasoning ability of LLMs, we conduct an in-depth analysis of the outcomes of IR and DR. We observe that the performance enhancement of DR is predominantly ascribed to two key aspects. Firstly, IR is effective in solving certain challenging problems that DR struggles with. Secondly, IR contributes to diversifying the reasoning process. The role of IR in these two scenarios is explicated through the following case studies.

\begin{figure}[t]
  \centering
  \includegraphics[width=\linewidth]{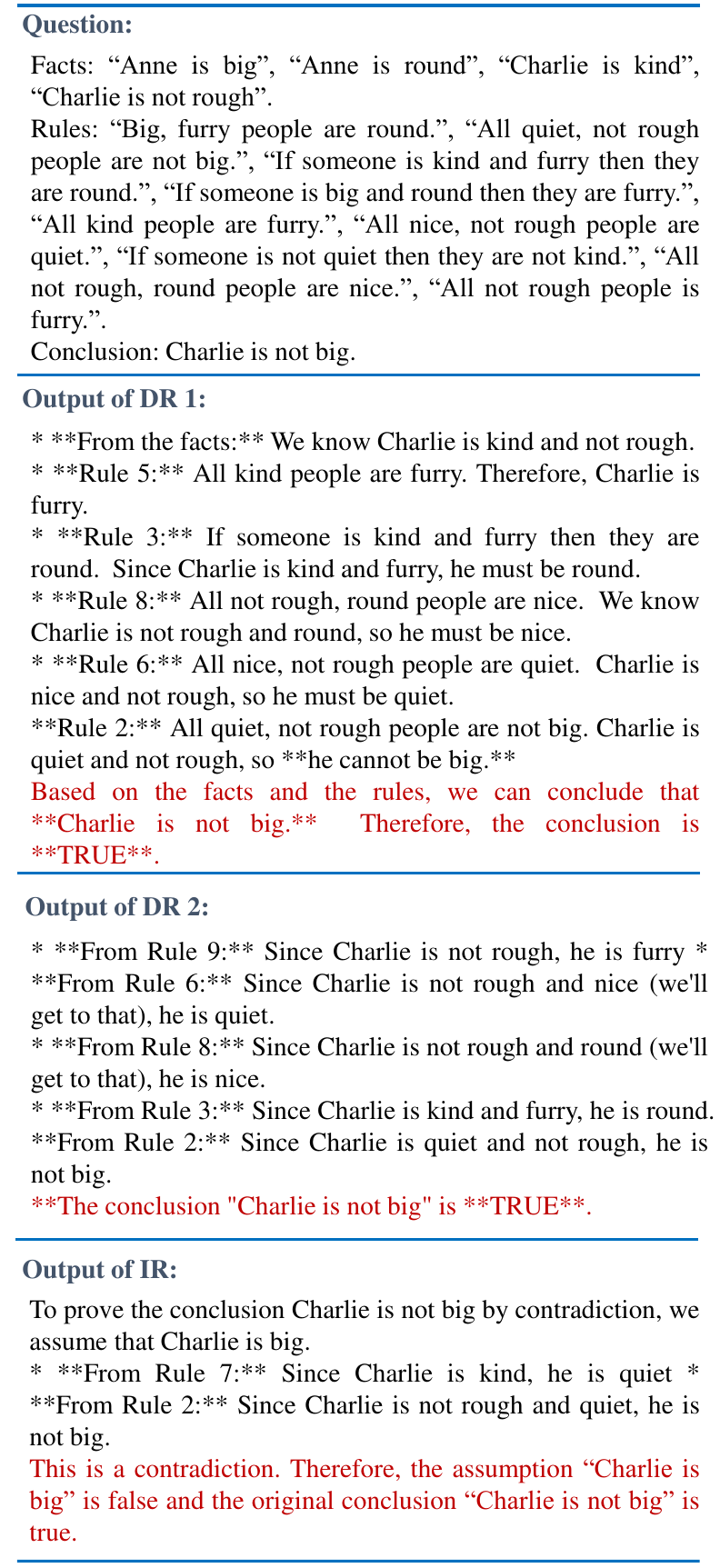}
  \caption{An example in ProofWriter demonstrates that DIR enriches the diversity of the reasoning paths of LLMs.}
  \label{fig:figA13}
\end{figure}
In Figure~\ref{fig:figA11}, it can be observed that the LLM lacks the ability to deduce the veracity or falsity of the issue, so it can only be judged as unknown. Through IR approach, which involves affirming the facts and rules while negating the conclusion, a contradiction can be derived. Consequently, it can be demonstrated that the negation of the conclusion is false, thereby validating the truth of the conclusion. Furthermore, Figure~\ref{fig:figA12} depicts a situation where DR yields incorrect proof, while IR is successful in solving a mathematic problem.

In Figure~\ref{fig:figA13}, it is evident that IR can offer a wider variety of reasoning paths. The analysis of the reasoning process reveals that multiple reasoning paths sampled from the LLM (\textit{i.e.}, ``Output of DR 1” and ``Output of DR 2”) yield similar reasoning paths, with discrepancies primarily in the selection of facts and rules during the reasoning process. However, the reasoning paths resulting from IR, which commence with the negation of the conclusion and identify contradictions with given facts and rules, differ significantly from the process of DR. 
\begin{figure}[t]
  \centering
  \includegraphics[width=\linewidth]{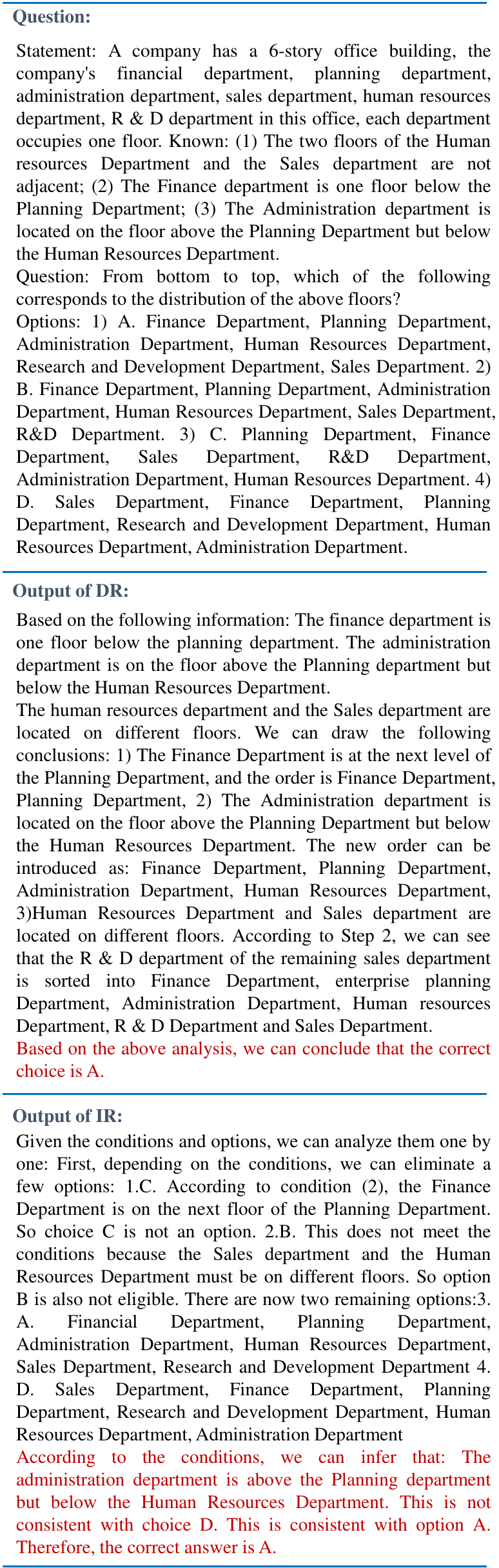}
  \caption{An example in LogiQA demonstrates that DIR enriches the diversity of the reasoning paths of LLMs.}
  \label{fig:figA14}
\end{figure}

IR can stimulate LLMs to generate more diverse reasoning paths. As stated in~\cite{evans2010intuition}, when a problem requires more deliberate thinking and analysis, the diversity of reasoning paths that can lead to the answer also increases. This ultimately helps to enhance the performance of LLMs. Figure~\ref{fig:figA14} illustrates an additional instance in which DIR enriches the diversity of the reasoning paths of LLMs.

\section{Description of Manual Validation}
In our experimental study, there is a requirement for human experts to confirm specific results for reliability. To ensure the integrity and consistency of the confirmation process, we have provided comprehensive training for the evaluators and established uniform evaluation criteria. The details are as follows:
\begin{itemize}
\item We have chosen 5 graduate students as evaluation experts to validate the experimental results. This group consists of 3 graduate students majoring in computer science and 2 graduate students majoring in mathematics. These experts possess strong logical reasoning abilities and mathematic expertise. We have also provided them with training in logical reasoning and mathematical proof tasks, along with a variety of cases to help them understand the requirements for these tasks.
\item To ensure the objectivity and fairness of the evaluation of the reasoning process, standardized evaluation criteria have been formulated for various tasks. A correct reasoning process does not contain any incorrect or omitted steps. Furthermore, for ProofWriter dataset, it is stipulated that the reasoning process should exclusively rely on the given facts and rules, without incorporating external or common knowledge.
\item Throughout the evaluation process, we maintain the concealment of the methods employed in each reasoning process from the evaluation expert. All data undergo random scrambling before being transmitted to the evaluation expert. Each evaluation expert conducts their assessment independently, and the final evaluation outcome is determined by the majority consensus of the assessments provided by multiple experts.
\end{itemize}

\section{Prompt Templates}\label{apx:C}
\begin{figure*}[t]
\begin{center}
\includegraphics[width=0.95\textwidth]{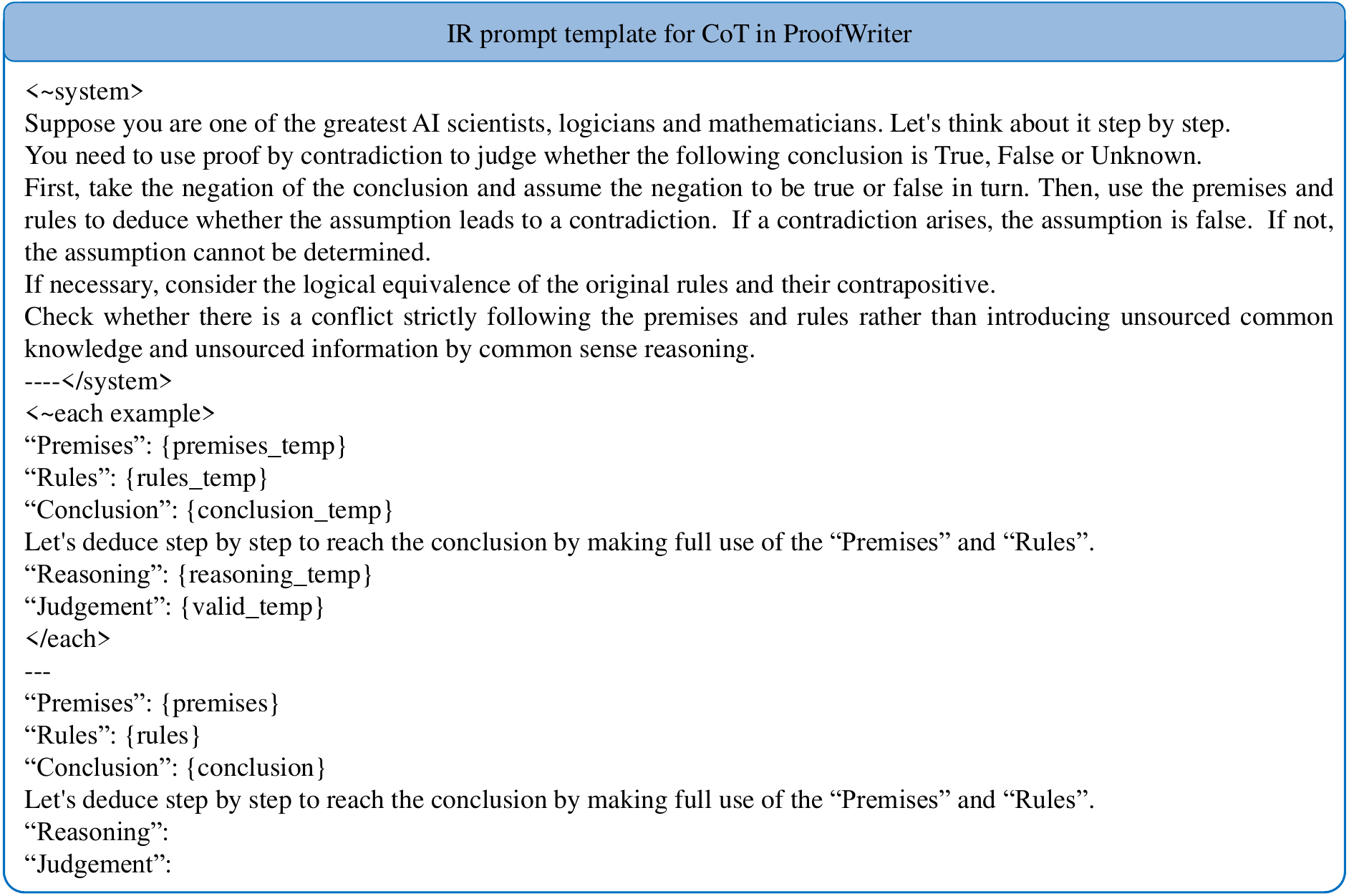}
\caption{IR prompt template for CoT in ProofWriter.}
\label{fig:figA1}
\end{center}
\end{figure*}

\begin{figure*}[t]
\begin{center}
\includegraphics[width=0.95\textwidth]{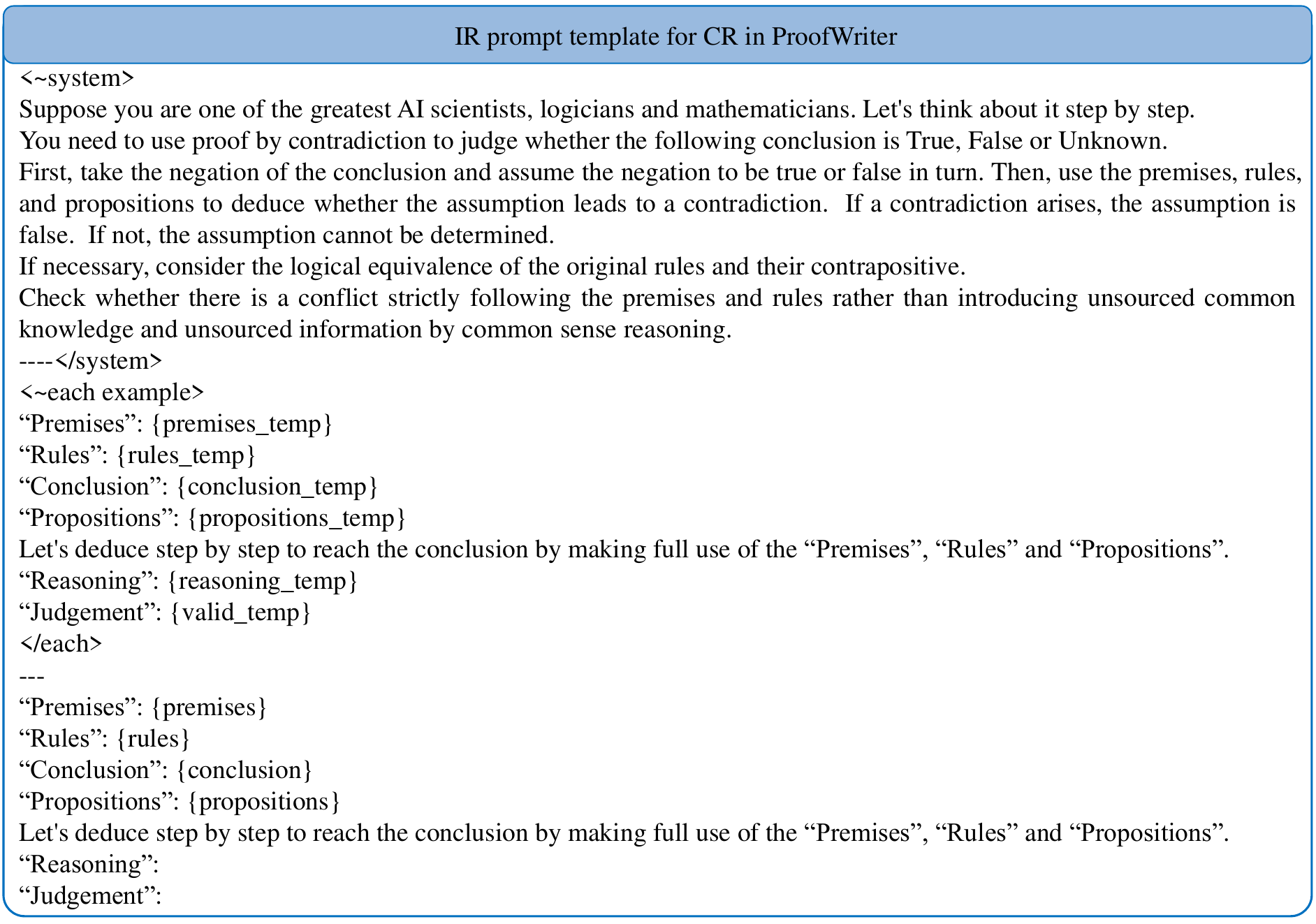}
\caption{IR prompt template for CR in ProofWriter.}
\label{fig:figA2}
\end{center}
\end{figure*}

\begin{figure*}[t]
\begin{center}
\includegraphics[width=0.95\textwidth]{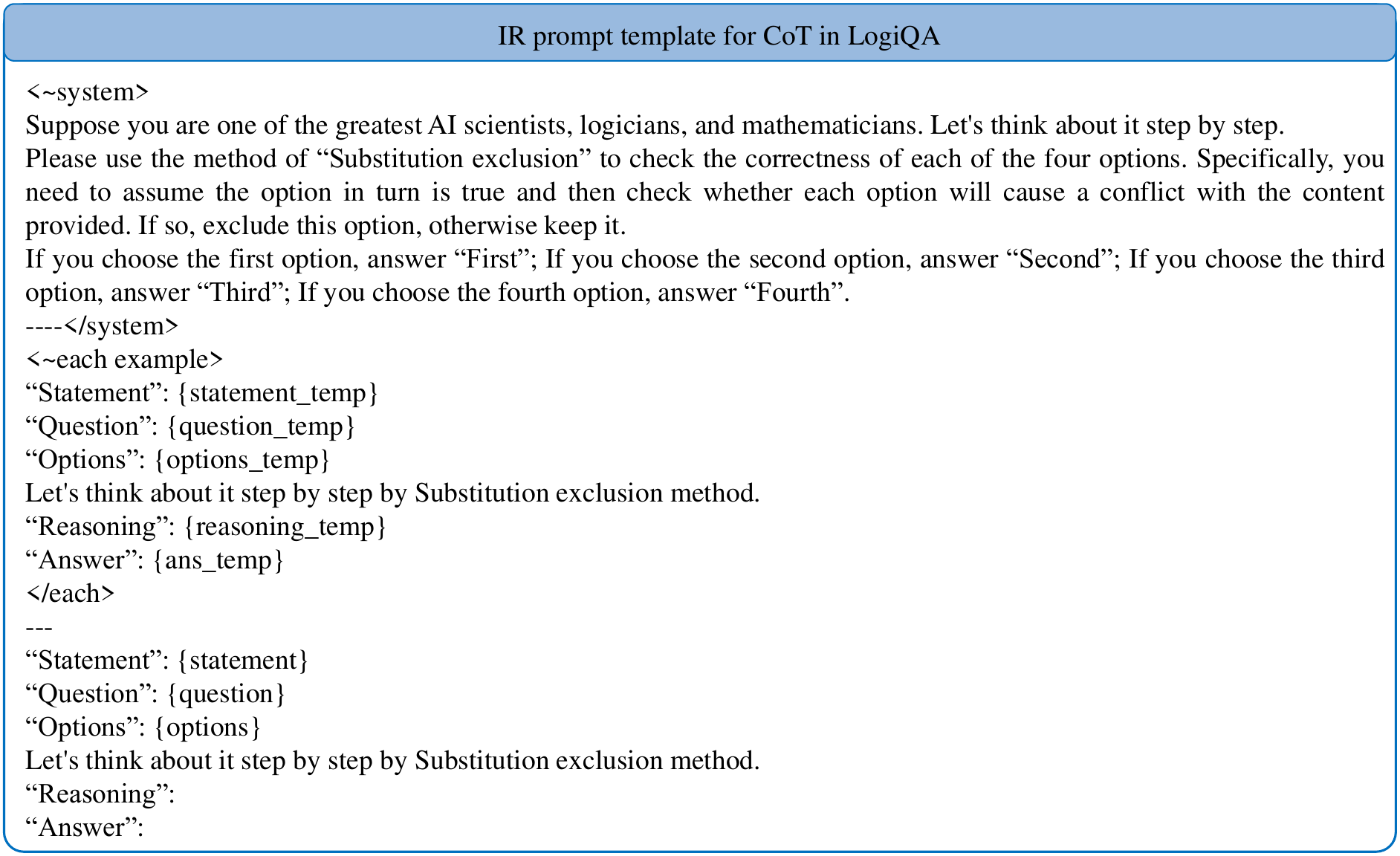}
\caption{IR prompt template for CoT in LogiQA.}
\label{fig:figA3}
\end{center}
\end{figure*}

\begin{figure*}[t]
\begin{center}
\includegraphics[width=0.95\textwidth]{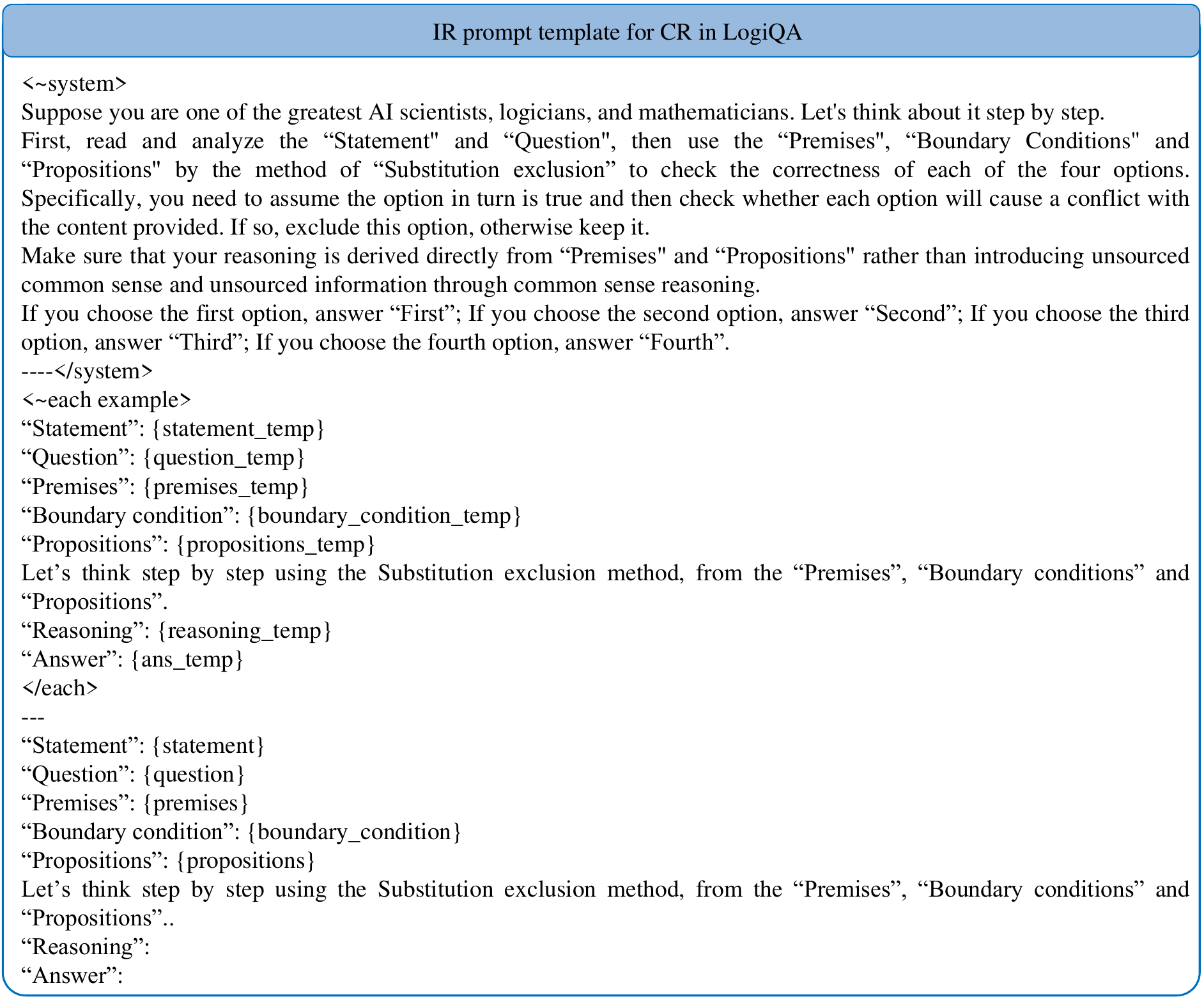}
\caption{IR prompt template for CR in LogiQA.}
\label{fig:figA4}
\end{center}
\end{figure*}

\begin{figure*}[t]
\begin{center}
\includegraphics[width=0.95\textwidth]{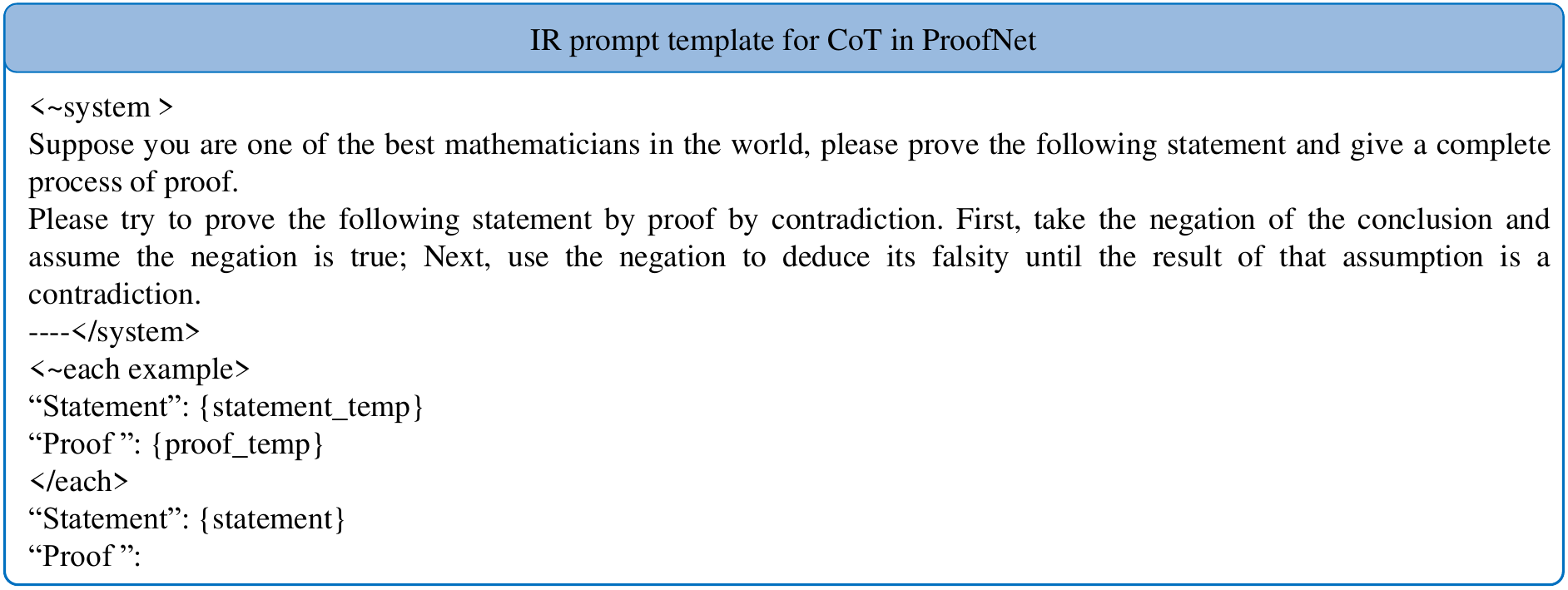}
\caption{IR prompt template for CoT in ProofNet.}
\label{fig:figA5}
\end{center}
\end{figure*}

\begin{figure*}[t]
\begin{center}
\includegraphics[width=0.95\textwidth]{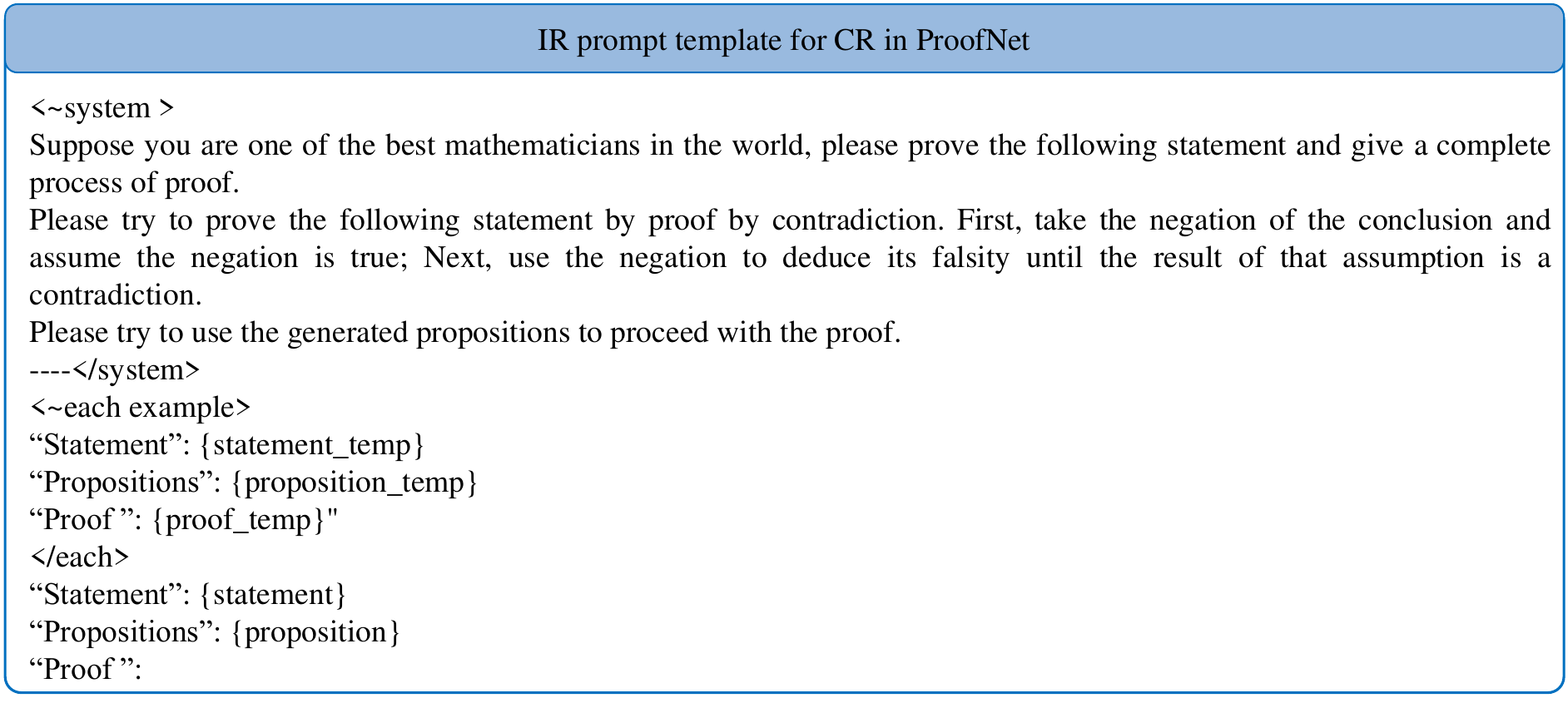}
\caption{IR prompt template for CR in ProofNet.}
\label{fig:figA6}
\end{center}
\end{figure*}

\begin{figure*}[t]
\begin{center}
\includegraphics[width=0.95\textwidth]{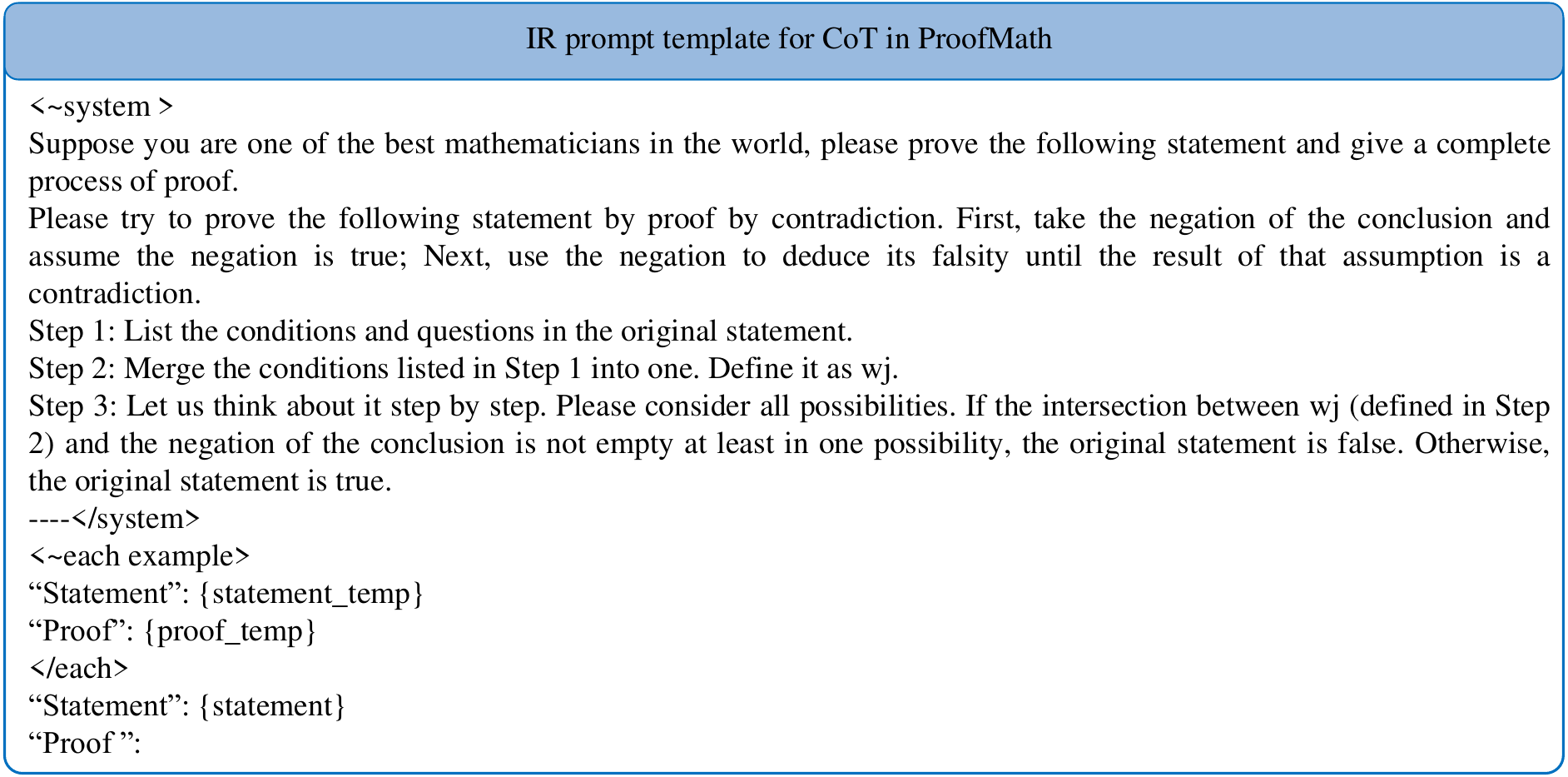}
\caption{IR prompt template for CoT in ProofMath.}
\label{fig:figA7}
\end{center}
\end{figure*}

\begin{figure*}[t]
\begin{center}
\includegraphics[width=0.95\textwidth]{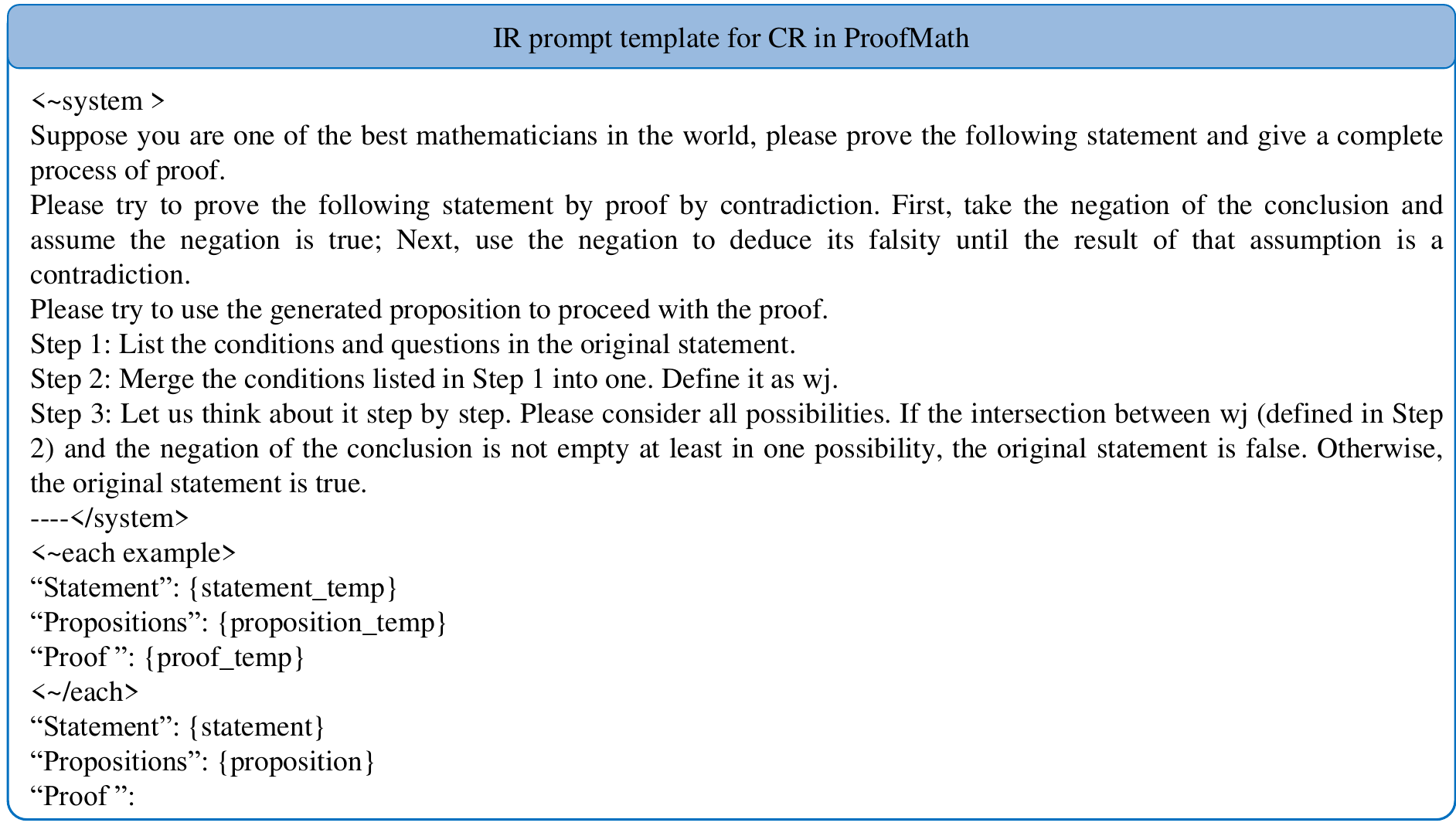}
\caption{IR prompt template for CR in ProofMath.}
\label{fig:figA8}
\end{center}
\end{figure*}

 Figures~\ref{fig:figA1}, \ref{fig:figA2}, \ref{fig:figA3}, \ref{fig:figA4}, \ref{fig:figA5},  \ref{fig:figA6}, \ref{fig:figA7}, and \ref{fig:figA8} illustrate the prompt templates employed in the experiment for different tasks. These templates primarily comprise IR instructions and examples demonstrating intermediate processes of IR.


\end{document}